\documentclass[sigconf]{acmart}

\AtBeginDocument{%
  }

\usepackage{hyperref}
\usepackage{url}
\usepackage{microtype}
\usepackage{graphicx} %
\usepackage{natbib}  %
\usepackage{caption} %
\usepackage{algorithm}
\usepackage{listings}
\usepackage{algorithmic}
\usepackage{amsmath}
\usepackage{booktabs}
\usepackage{multirow}
\usepackage[outdir=./]{epstopdf}
\usepackage{enumitem}
\usepackage{caption}
\usepackage{subcaption}
\usepackage{subfloat}
\usepackage{newfloat}
\usepackage{graphicx}
\usepackage{svg} %
\usepackage[normalem]{ulem}
\usepackage{framed}
\usepackage{mdframed}
\usepackage{xcolor}
\usepackage{lipsum}
\usepackage{float}
\usepackage{hyperref}
\usepackage{tabularx}
\usepackage{wrapfig}
\usepackage{bm}
\usepackage{colortbl}
\usepackage{amsthm}

\usepackage{balance}

\definecolor{shadecolor}{gray}{0.9}

\usepackage{tcolorbox} 
\tcbuselibrary{theorems} 
\usepackage{xcolor} 
\definecolor{mytheoremfr}{RGB}{200,200,200} 
\definecolor{mytheorembg}{RGB}{240,240,240} 

\newtheorem*{remark}{Remark}

\newtcbtheorem{Prompt}{Prompt}{colback=mytheorembg!25,colframe=mytheoremfr, boxrule=0.5pt}{th}

\setcopyright{acmlicensed}
\acmDOI{10.1145/3746027.3755873}
 \acmISBN{979-8-4007-2035-2/2025/10}

\copyrightyear{2025}
\acmYear{2025}
\setcopyright{acmlicensed}\acmConference[MM'25]{Proceedings of the 33th ACM International Conference on Multimedia}{October 27--31, 2025}{Dublin, Ireland}
\acmBooktitle{Proceedings of the 33th ACM International Conference on Multimedia (MM '25), October 27--31, 2025, Dublin, Ireland}

\begin{document}

\title{TV-RAG: A Temporal-aware and Semantic Entropy-Weighted Framework for Long Video Retrieval and Understanding}

\author{Zongsheng Cao}
\authornote{Equal contribution}
\email{agiczsr@gmail.com}
\affiliation{
  \institution{Researcher}
  \country{}
}

\author{Yangfan He}
\authornotemark[1]
\email{he00577@umn.edu}
\affiliation{
  \institution{UMN 
  }
  \country{}
  }
  
\author{Anran Liu}
\authornotemark[1]
\authornote{Corresponding author}
\email{anniegogo1008@gmail.com}
\affiliation{
  \institution{Researcher}
  \country{}
}

  \author{Feng Chen}
 \email{chenfeng@lenovo.com}
 \affiliation{%
   \institution{PCIE}
   \country{}
 }

  \author{Zepeng Wang}
  \email{wangzpb@lenovo.com}
  \affiliation{%
    \institution{PCIE}
    \country{}
  }

    \author{Jun Xie} 
    \authornotemark[2]
  \email{xiejun@lenovo.com}
  \affiliation{
    \institution{PCIE}
    \country{}
  }

\renewcommand{\shortauthors}{Zongsheng Cao et al.}

\begin{abstract}
  Large Video Language Models~(LVLMs) have rapidly emerged as the focus of multimedia AI research. Nonetheless, when confronted with lengthy videos, these models struggle: their temporal windows are narrow, and they fail to notice fine-grained semantic shifts that unfold over extended durations. Moreover, mainstream text-based retrieval pipelines, which rely chiefly on surface-level lexical overlap, ignore the rich temporal interdependence among visual, audio, and subtitle channels. To mitigate these limitations, we propose TV-RAG, a training-free architecture that couples temporal alignment with entropy-guided semantics to improve long-video reasoning. The framework contributes two main mechanisms:  
  \emph{(i)} a time-decay retrieval module that injects explicit temporal offsets into the similarity computation, thereby ranking text queries according to their true multimedia context; and  
  \emph{(ii)} an entropy-weighted key-frame sampler that selects evenly spaced, information-dense frames, reducing redundancy while preserving representativeness. 
  By weaving these temporal and semantic signals together, TV-RAG realises a dual-level reasoning routine that can be grafted onto any LVLM without re-training or fine-tuning. The resulting system offers a lightweight, budget-friendly upgrade path and consistently surpasses most leading baselines across established long-video benchmarks such as Video-MME, MLVU, and LongVideoBench, confirming the effectiveness of our model. The code can be found at https://github.com/AI-Researcher-Team/TV-RAG.
\end{abstract}





\maketitle

\section{Introduction}
\label{sec:intro}

Recent breakthroughs in large-scale language modelling have catalysed rapid progress in multimodal research, ultimately leading to a new class of Large Video–Language Models~(LVLMs)~\cite{ videollama,  liu2024oryx, li2025llama}. Despite their impressive accuracy on short, clip-length inputs, current LVLMs still face significant obstacles when tasked with analyzing and reasoning over very long videos.

Recent attempts to improve long-horizon reasoning in Large Video Language Models, exemplified by LongVA \cite{longva} and LongLLAVA \cite{longllava}, have largely focused on widening the context window. LongVA, for instance, simply scales up the token capacity to ingest more frames. This brute-force expansion, however, proves brittle when faced with out-of-distribution footage: the Video-MME benchmark shows its accuracy drops sharply as additional frames are supplied.  A parallel research thread employs Retrieval-Augmented Generation (RAG) to supplement video queries with externally retrieved documents~\cite{llavanextvideo, luo2024video, llavanext}. Yet, treating the video stream purely as text ignores vital visual alignment cues, and the reliance on high-capacity LVLM backbones translates into steep computational costs and poor adaptability. Consequently, existing RAG pipelines struggle to capture the intertwined temporal and semantic signals required for truly robust long-video understanding.

Based on the considerations above, we focus on the following question:
\textit{Can we develop a new model that can tackle the video RAG task from both temporal and semantic perspectives in a unified training-free framework?}

Motivated by this limitation, several studies~\cite{llavanext, luo2024video, llavanextvideo} advocate replacing the bulky chain of visual tokens with proxy captions distilled directly from the video stream via off-the-shelf optical character recognitions (OCR), automatic speech recognition (ASR), and object-detection models. These auxiliary texts are tightly coupled to the visual scene and inject complementary cues unavailable in raw pixels. For instance, Long-context LVLMs extend visual tokens so the model sees every frame, yielding strong temporal awareness but at the cost of heavy retraining and limited semantic depth.  GPT-based agent pipelines first convert videos into textual reports, then query a proprietary LLM such as GPT-4o; while semantically powerful, they lose fine-grained temporal cues and depend on closed-source services. Yet, the strategy leaves two key issues unresolved:
\begin{itemize}
    \item[\textbf{C1}] Cross-modal dependencies among images, audio, and subtitles remain implicit, hampering performance on tasks that demand joint reasoning over long temporal spans;
    \item[\textbf{C2}] The fixed context window still cannot follow subtle semantic drifts that emerge gradually throughout lengthy footage.
\end{itemize}

\begin{figure}[t!]
  \centering 
  \includegraphics[width=1.0\linewidth]{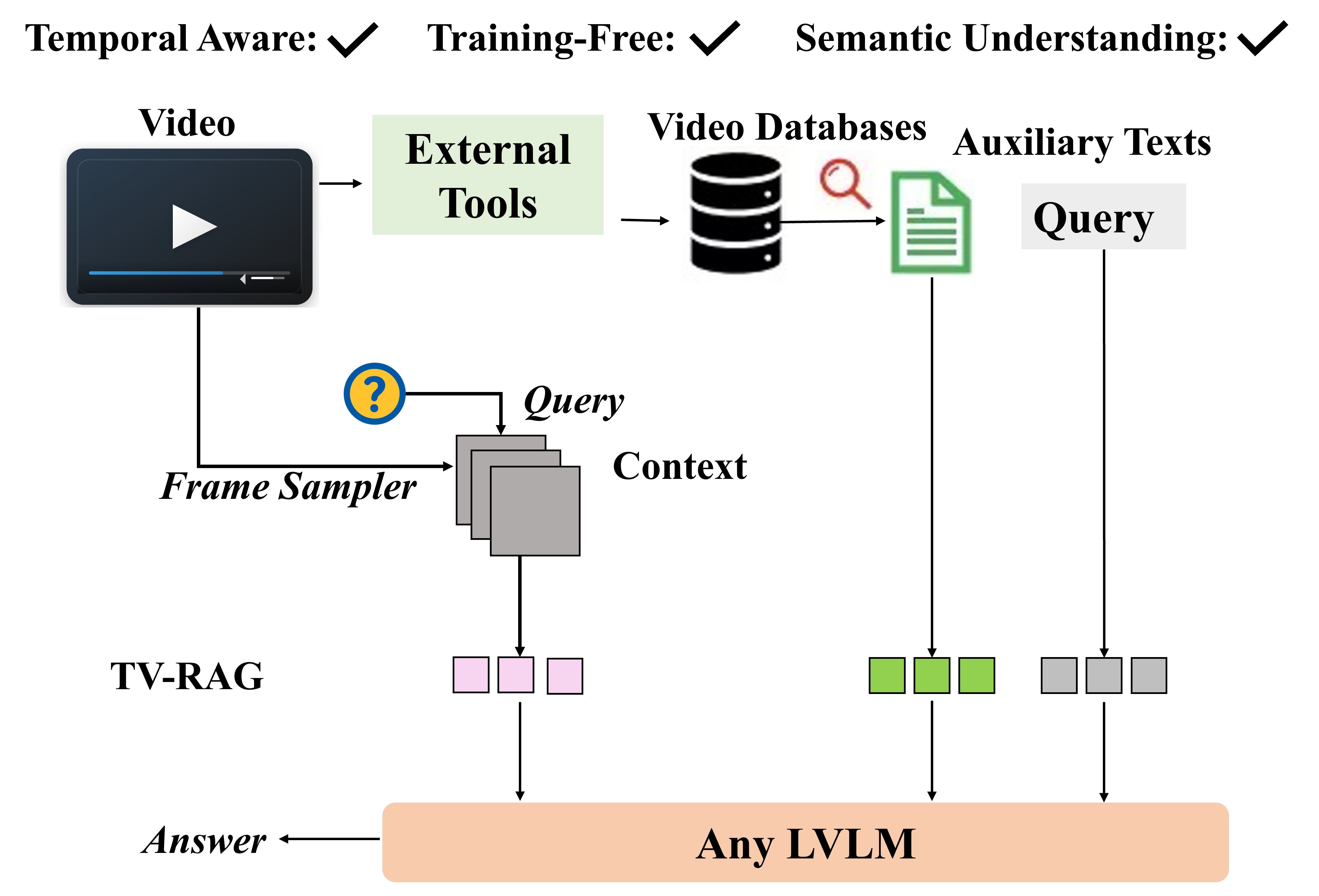}
  \caption{Advantages of our TV-RAG. TV-RAG provides a temporal-aware, semantic-aware, and training-free pipeline that is easily compatible with any LVLM.}
  \label{fig_intro} 
\end{figure}

To this end, we introduce TV-RAG, a novel framework designed to enhance the comprehension of long videos in LVLMs without requiring additional training.  Specifically, to address (C1), TV-RAG employs a semantic entropy-based weighting strategy for key frame selection to evenly distribute selected frames across time, reducing redundancy, enhancing representativeness, and prioritizing the most informative frames.  To tackle (C2), it incorporates a temporal window-based BM25 model that integrates time-aligned auxiliary data, including OCR, ASR, and object detection (DET) outcomes. This approach enhances the relevance of text queries by aligning them with the temporal context of multimedia content, ensuring that retrieved information is contextually appropriate.

By combining explicit temporal order with entropy-based semantic salience, TV-RAG enables a two-stage reasoning process that removes noise and filters out irrelevant content from long videos. The module is lightweight and can be added to any existing LVLM without modifying its weights. This plug-and-play design allows TV-RAG to outperform popular models on benchmark datasets such as Video-MME, MLVU, and LongVideoBench. We further evaluate TV-RAG on six widely used open-source LVLM backbones, confirming its effectiveness.

In summary, our contributions are three-fold:

\begin{itemize} \item To mitigate frame redundancy and preserve cross-modal semantics in long videos, we propose TV-RAG, including a semantic-entropy key-frame selector that computes multimodal information gain and enforces uniform temporal coverage, yielding compact yet representative frame sets for downstream LVLM reasoning.

  \item To handle semantic drift and maintain temporally coherent retrieval, we introduce a temporal-window BM25 retriever that ties lexical relevance to timestamp alignment across auxiliary text streams, supplying the LVLM with context that stays synchronized with the evolving video content, all without modifying model weights.
   
  \item Extensive experiments on several well-established benchmarks show that TV-RAG achieves state-of-the-art performance and outperforms other models. It can also be used as a plug-in for other models. \end{itemize}

\section{Related Work}
\label{sec:related_work}

\begin{figure*}[t!]
  \centering 
  \includegraphics[width=1.0\linewidth]{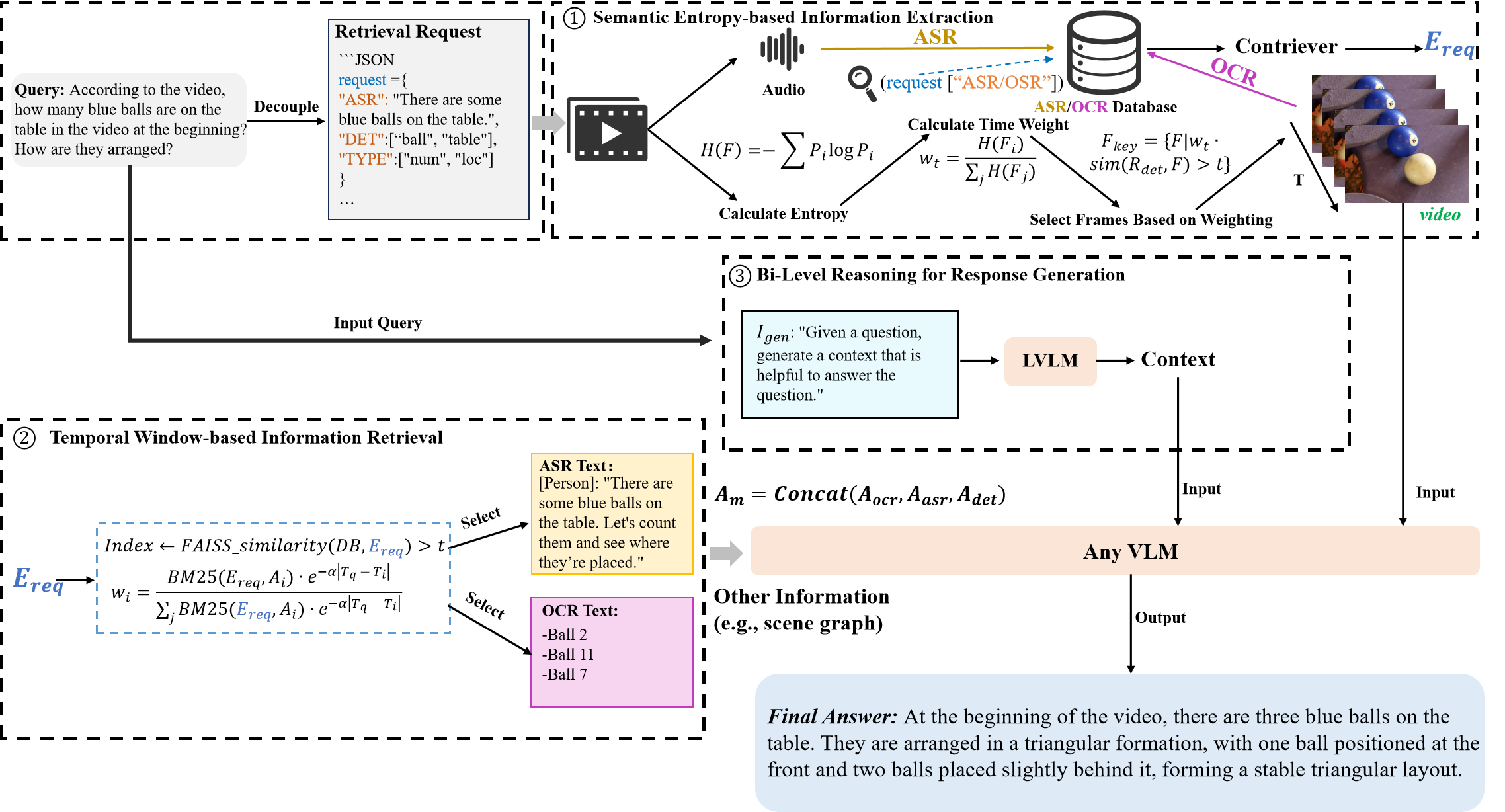}
  \caption{The illustration of our framework TV-RAG. The pipeline begins with query decoupling, where the LVLM rewrites the user question into an explicit evidence-retrieval request, cleanly separating search from reasoning. TV-RAG then executes three tightly coupled steps: (i) a semantic-entropy selector distills high-information frames across visual, OCR, ASR and detection streams; (ii) a temporal-decay BM25 retriever ranks auxiliary texts by both lexical relevance and timestamp proximity, ensuring chronological coherence; and (iii) a bi-level reasoning routine drafts and self-verifies the answer using the retrieved evidence. By closing the loop between asking, retrieving, and reasoning without altering LVLM weights, TV-RAG simultaneously sharpens retrieval precision and elevates answer faithfulness.}
  \label{fig_framework} 
\end{figure*}

\subsection{Large Language Models for Video}

The surge of large language models has sparked parallel efforts to craft versatile video–language systems.  Early work, Video-ChatGPT \cite{videochatgpt}, processes frames separately and later fuses their representations via spatial–temporal pooling.  VideoChat~\cite{videochat} augments appearance embeddings with on-the-fly textual captions to form a richer clip encoding.  To minimise the gap between image and video pathways, Video-LLaVA~\cite{videollava} introduces a shared projector that aligns both outputs within a common language latent space.  Its successor, LLaVA-NeXT-Video~\cite{llavanextvideo}, adapts the LLaVA-NeXT backbone~\cite{llavanext} through targeted video fine-tuning.  Despite these advances, retrieval-augmented schemes such as Video-RAG~\cite{luo2024video} still fall short in tracking the subtle temporal dependencies and layered semantics present in lengthy, information-dense footage.

\subsection{Large Language Models for Long-context Video}

A prominent strand of recent work attempts to enlarge the effective context window so that models can reason over intricate video narratives.  LongVA \cite{longva} and Long-LLaVA \cite{longllava} pursue this goal by first pre-training language backbones on massive text corpora, banking on the resulting long-sequence handling skills to transfer to video tasks.  In contrast, INTP \cite{intp} rearranges the incoming video tokens without any additional training, stretches the window of an LVLM to ingest far more visual information.  Yet, these solutions confront a familiar trade-off: every extra batch of sampled frames inflates computation while delivering only marginal gains.  Because videos contain significant redundancy, and model capacity is finite. Under these circumstances, most models cannot achieve satisfactory performance.

\subsection{Video Understanding by GPT-based Agent}

A complementary line of work investigates using an LLM as a controller that calls external vision or audio modules to parse long-form video data, especially for question–answering tasks~\cite{min2024morevqa, zhang2023simple, wang2024videoagent, gupta2023visual, suris2023vipergpt}. MM-VID \cite{mmvid} advances this idea by explicitly pairing each frame with its textual caption, whereas VLog \cite{vlog} first extracts audio and appearance cues via multimodal pre-training and then distills them into compact textual surrogates.  Agent-style frameworks such as VideoAgent \cite{videoagent}, DrVideo \cite{drvideo}, and OmAgent \cite{omagent} go a step further, issuing adaptive queries to retrieve relevant segments before reasoning over them.  Despite their ingenuity, these pipelines still suffer from two practical drawbacks: (i) iterative tool invocation inflates inference latency, and (ii) many rely on closed-source models, limiting both efficiency and the ease with which the community can replicate results using purely open-source stacks. Moreover, there are also some advanced methods, such as \cite{luo2024video}; however, they have flaws in capturing the joint associations of temporal and semantic information. In this way, it is still an open issue to be addressed.

\section{Methodology}
\label{sec:formatting}

We introduce TV-RAG, a novel training-free process designed for LVLMs that can be seamlessly integrated into any existing LVLM. As shown in Fig. \ref{fig_framework}, the process consists of three main phases: \textbf{(\romannumeral 1) Semantic entropy-based information extraction:} After obtaining the query, information is extracted based on semantic entropy from different sources. \textbf{(\romannumeral 2) Temporal decay-enhanced retrieval model:} In order to capture the important temporal information in the video, the time window mechanism request is introduced for obtaining the relevant information. \textbf{(\romannumeral 3) Context-enhanced reasoning-based response generation:} In this final stage, the auxiliary text retrieved based on the context-enhanced reasoning mechanism is integrated with the user's query and fed into the LVLM to generate the final output.

\noindent\textbf{Problem Setup. } Let \( \mathcal{V} \) be an input video.  We then use a frame–selection unit to extract \(N\) representative images \( \mathcal{F} \)
Each frame is then mapped into a visual embedding via a frozen image encoder, e.g., \ CLIP-L~\cite{clip}, yielding  
\( \mathcal{F}_{\!v}\) from \( \mathcal{F} \).  
Finally, the visual tokens \( \mathcal{F}_{\!v} \) and a user query \( \mathcal{Q} \) are supplied to a large video–language model to generate the answer \( \mathcal{O} \):
\begin{equation}
    \mathcal{O}= \operatorname{LVLM}(\mathcal{F}_{\!v}, \mathcal{Q}).
\end{equation}
In this way, we complete the RAG process for videos.

\noindent\textbf{Two-stage Processes}. In this paper, motivated by previous efforts \cite{videollava,luo2024video}, upon receiving a user's query regarding a video, the LVLM follows a two-phase process. First, the LVLM decouples the query to generate structured retrieval requests, which serve as auxiliary inputs for later stages. In this phase, the LVLM focuses solely on processing textual data without access to the video frames. These requests are then parsed, and if any category does not require data, it will be marked as NULL (indicating no need for that specific request). This results in the structured retrieval requests, \(\bm{\mathrm{R}} = \{\bm{\mathrm{R}}_{i}\}\), which means the request categories, such as detection information. Then, they are passed on to the next phase. The process can thus be decoupled into two main stages, as formalized in the following equations:
\begin{equation}
    \bm{\mathrm{R}} = \texttt{LVLM}(\bm{\mathrm{P}}, \bm{\mathrm{Q}}), \quad \bm{\mathrm{R}} = \{\bm{\mathrm{R}}_{i}\},
\end{equation}
where $\bm{\mathrm{P}}$ is the prompt.
In the second phase, these requests, \(\bm{\mathrm{R}}\), along with the video frames, \(\bm{\mathrm{F}_v}\), are used to generate the final output:
\begin{equation}
    \bm{\mathrm{O}} = \texttt{LVLM}(\bm{\mathrm{F}_v}, \bm{\mathrm{Q}}, \bm{\mathrm{R}}).
\end{equation}
In this phase, the LVLM processes the video frames, utilizes the textual query, and incorporates the generated retrieval requests to provide a more informed and contextually relevant answer.

\subsection{Semantic Entropy-based Extraction}

\noindent\textbf{Auxiliary‐Text Extraction and Retrieval.}
Following the blueprint laid out by recent retrieval-augmented systems~\cite{llavanextvideo, luo2024video, llavanext}, we first transcribe the raw video into several complementary text channels and then fetch the portions most useful for the downstream LVLM.  
Formally, we create a set  
\(
\mathcal{R}= \{\, \mathcal{R}_{asr},\, \mathcal{R}_{ocr},\, \mathcal{R}_{det} \,\},
\)  
where  
(\textit{i}) \(r_{\textsc{asr}}\) is automatic speech recognition to harvest dialogue or soundtrack content;  
(\textit{ii}) \(r_{\textsc{det}}\) is salient entities present on screen; and  
(\textit{iii}) \(r_{\textsc{ocr}}\) is optical character recognition to capture any scene text.  

Longer clips naturally produce voluminous and often redundant token streams, while the context budget of current open-source LVLMs remains tight.  To avoid overflow, we retrieve only those snippets that align semantically with the user query.  
Before retrieval, three modality-specific repositories are built in parallel from the video itself: an ASR base \(\mathcal{D}_{\textsc{asr}}\), an OCR base \(\mathcal{D}_{\textsc{ocr}}\), and an object-detection base \(\mathcal{D}_{\textsc{det}}\).  Subsequent look-ups are executed against these lightweight databases, ensuring that irrelevant tokens do not enter the LVLM’s limited context window.

\noindent\textbf{Building Retrieval Repositories.}
Open source LVLMs tend to misread scene text and spoken words, falling short of their proprietary counterparts.  
To curb such hallucinations and exploit frame content more effectively, we offload text extraction to specialist models.  
Concretely, EasyOCR~\cite{easyocr} is run on each key frame to harvest on-screen captions, giving a pool of strings \(\mathbf{T}_{\textsc{ocr}}\);
meanwhile, the soundtrack is transcribed by Whisper ~\cite{whisper}, yielding an ASR transcript \(\mathbf{T}_{\textsc{asr}}\) as advocated in prior work~\cite{llavanextvideo, luo2024video, llavanext}.  
Both text streams are embedded with the \textsc{Contriever} encoder~\cite{contriever} to obtain dense vectors, which are written to two separate FAISS indices~\cite{faiss}:  
\(\mathcal{D}_{\textsc{ocr}}\) for scene text and \(\mathcal{D}_{\textsc{asr}}\) for speech.  
This design enables low-latency, similarity-based retrieval of the most relevant snippets during query time.

\noindent\textbf{Key–Frame Selection.}  
Although modern LVLMs excel at recognising objects, they remain error-prone when asked to count instances, pinpoint locations, or reason about complex interactions, frequently hallucinating details when contextual cues are sparse.  
To curb this issue, motivated by \cite{videollava,luo2024video}, we rank every sampled frame \(\mathcal{F}=\{F_t\}\) by the semantic affinity between the detector request \(R_{det}\) and the frame content, modulated by a temporal importance weight:  
\[
{F}_{key}
   = \Bigl\{F_t \,\bigm|\,
          \alpha_t \cdot 
          \operatorname{CLIP}\bigl(R_{det},F_t\bigr)
          \ge \tau\Bigr\},
\]
where \(\tau\) is a similarity threshold.  
The weight \(\alpha_t\) captures how much new information the segment contributes and is computed from the normalised Shannon entropy of its visual features:  
\[
\alpha_t \;=\;
\frac{H(F_t)}{\sum_{j} H(F_j)}.
\]  
Object detection is then run only on this entropy-aware subset \(\mathcal{F}_{\textsc{key}}\), ensuring that the LVLM processes the most informative and contextually relevant frames.

The weight \( w_t \) assigned to each frame is determined by the \textit{information entropy} \( H(F_t) \), a measure of the uncertainty or variability within the frame’s content. Then the entropy \( H(F_t) \) is computed as:
\begin{equation}
    H(F_t) = - p_t \log p_t, \quad p_t = \frac{\text{CLIP similarity}(R_{\text{det}}, F_t)}{\sum_j \text{CLIP similarity}(R_{\text{det}}, F_j)},
\end{equation}
where \( p_t \) represents the normalized similarity between the object retrieval query and each frame \( F_t \), and \( H(F_t) \) captures how much variance exists in the content of a given time segment. The selection process begins by computing the CLIP similarity \cite{clip} between an object retrieval query \( \bm{\mathrm{R}}_{det} \) and the sampled video frames \( \bm{\mathrm{F}} \).  High entropy values indicate significant changes in the visual or auditory content, suggesting that this segment contains crucial, non-redundant information.

\begin{remark}
    In this way, it ensures that time segments with higher variability in content are more likely to be selected, while segments with low variability (i.e., those that are redundant) are deprioritized. The goal is to maximize the representativeness of the selected frames and minimize redundant or irrelevant content, which is particularly important in the context of long-duration videos. By incorporating this entropy-based weighting mechanism, we ensure that critical moments in the video, which may involve sudden shifts in object interaction or scene changes, are not overlooked. 
\end{remark}

\subsection{Temporal Window-based Text Retrieval}

\noindent\textbf{Query–Aware Retrieval.}  
Our pipeline introduces a dedicated retrieval routine that tightly couples the user prompt with the auxiliary captions mined from the video.  
Inspired by prior multimodal RAG work~\cite{videollava, luo2024video}, we begin by converting the query into a dense representation \(\mathbf{e}_{req}\) using the Contriever encoder.  

To prevent the common temporal drift problem, where a lexically relevant caption is drawn from an incorrect moment in the clip, we refine the top-\(k\) text hits with a temporal re-ranking stage.  
Each candidate's score is updated by a weighted sum of its lexical similarity to \(\mathbf{e}_{req}\) and its time-stamp distance from the query segment, ensuring the final selection is both semantically pertinent and temporally on point.

\noindent\textbf{Temporal Re-scoring of Retrieved Snippets.}
Assume that a FAISS search over the OCR/ASR indices yields the relevant snippets
\(\mathcal{C}\).
Let \(\tau_q\in\mathbb{R}\) mark the moment in the video that anchors the user's query, and denote by \(\tau_i\) the original time-code of snippet \(c_i\). We refine each snippet's lexical BM25 score by a time-aware decay:
\begin{equation}
    \tilde{s}_i \;=\;
    \frac{\operatorname{BM25}(\mathbf{e}_{\text{req}},c_i)\,
          e^{-\sum_{k=0}^2\lambda_k|\tau_q^{k}-\tau_i|}}
         {\displaystyle
          \sum_{j=1}^{3K}\operatorname{BM25}(\mathbf{e}_{\text{req}},c_j)\,
          e^{-\sum_{k=0}^2\lambda_k|\tau_q^{k}-\tau_j|}},
\end{equation}
where \(\mathbf{e}_{\text{req}}\) is the Contriever embedding of the query and \(\lambda>0\) (set to \(1\) in our study) controls the strength of the temporal penalty. Here the three anchor points \(\tau_q^{k}\) serve complementary roles: \(\tau_q^{0}\) is the time-code of the \emph{last} video frame, \(\tau_q^{1}\) marks the \emph{first} frame, and \(\tau_q^{2}\) corresponds to the frame whose local context is semantically most similar to the query.  The rationale is that salient evidence can appear near a video's conclusion (e.g., a reveal or punch-line), at its outset (introductory setup), or at an unpredictable location that is best located via semantic affinity.  By injecting all three anchors into the exponential decay, the rescoring scheme remains agnostic to where the key information actually lies, yet still rewards snippets clustered around whichever temporal region proves relevant for the user's request.

The exponential term softly favours snippets that occur near \(\tau_q^k\) while still allowing distant, but lexically pertinent candidates to survive with a lower confidence.   Finally, we keep the top-\(K\) snippets with the highest
\(\tilde{s}_i\):
\begin{equation}
    \mathcal{C}\;=\;\texttt{TopK}\!\bigl(\mathcal{C},\,\tilde{s}_i\bigr).  
\end{equation}

Incorporating timestamp proximity into the similarity score remedies two chronic weaknesses of standard retrieval pipelines.  
(i) It sharply reduces the likelihood of selecting passages that look semantically relevant yet come from the wrong moment in the footage, a pitfall common to long videos with recurrent motifs.  
(ii) It maintains chronological consistency for the LVLM, so downstream reasoning and text generation remain anchored to the correct slice of the narrative.  
Unlike hard sliding windows or ad-hoc alignment rules, the proposed weighting scheme is continuous, differentiable, and plug-compatible with existing multimodal retrieval frameworks.

\noindent\textbf{Object Information Retrieval.} After retrieving the relevant information $\bm{\mathrm{A}}$, the next step involves extracting \textit{object-related information} stored in the auxiliary database. Motivated by \cite{llavanextvideo,luo2024video,llavanext}, object detection models typically produce raw outputs that include spatial attributes in the form of bounding boxes and other information.  These components collectively form a structured scene graph, expressed as $\mathrm{A}_{\mathrm{st}} $, which can be regarded as a detection database and enhances the ability of LVLMs to interpret temporal and spatial information. By incorporating this structured graph representation, we enable LVLMs to reason over the spatial layout and temporal dynamics of objects within the video.

\subsection{Context-augmented Reasoning with Query Reformulation}

While retrieved auxiliary signals (e.g., OCR, ASR) can provide essential external context, relying exclusively on these sources introduces two key limitations: (i) they are often noisy, fragmented, or partially misaligned with the user's underlying intent; and (ii) compressing them to fit the limited context window of LVLMs may introduce semantic loss and degrade the coherence of downstream reasoning.

To overcome these issues, we propose a context-augmented reasoning strategy that enhances retrieved evidence through a dual-channel mechanism. This approach not only supplements the auxiliary inputs but also enhances the model's understanding of the user query via semantic rephrasing.

\noindent\textbf{Level-1: Evidence-based Contextualization.}
We first aggregate auxiliary textual signals from $A_{st}$.
This aggregated context reflects the explicit knowledge retrieved from the video content and acts as grounded, observable evidence for reasoning.

\noindent\textbf{Level-2: Query-enhanced Latent Reasoning.}
Recognizing that retrieved evidence may still omit latent or abstract concepts implied by the user's query, we introduce a generation-based augmentation step. Instead of compressing auxiliary data, we instruct the LVLM to perform two complementary operations:
(1) generate general background context relevant to the query to enrich model-level knowledge grounding; (2) reformulate the user query into a set of semantically similar paraphrases to enhance retrieval sensitivity and decoding diversity.

Specifically, we define the prompt $I_{gen}$ to guide context generation as:
\begin{quote}
    \texttt{Given a question, first generate a helpful background context. Then, provide 2-3 alternative phrasings of the question with similar meaning.}
\end{quote}
This produces two outputs: a generated context $\bm{\mathrm{C}}_{\text{gen}}$ and a set of $k$ semantically equivalent queries $\{\bm{\mathrm{Q}}^{(i)}\}_{i=1}^k$. These components enable the model to reason over both retrieved facts and inferred intent.

\noindent\textbf{Final Input Composition.}
The final structured input to the LVLM integrates the sampled video frames, the retrieved auxiliary evidence, the original query, and the generated latent context:
\begin{equation}
\bm{\mathrm{O}} = \texttt{LVLM}(F_{key}, \texttt{Concat}(\bm{\mathrm{A}}_{st}, \bm{\mathrm{Q}}, \{\bm{\mathrm{Q}}^{(i)}\}_{i=1}^k, \bm{\mathrm{C}}_{\text{gen}})).
\end{equation}

\begin{remark}
    By combining retrieval-grounded evidence with generative reasoning and query reformulation, this bi-level design strengthens both explicit and implicit context understanding. It also mitigates alignment gaps between noisy auxiliary signals and the model's pre-trained knowledge space, resulting in more robust and semantically coherent outputs.
\end{remark}

\section{Experiments}
\label{sec:exp}
\begin{table*}[t!]
\centering

\label{tab_videomme}
\setlength{\tabcolsep}{2.5mm}
\renewcommand\arraystretch{0.95} 
\begin{tabular}{l|ccc|cccc|c}
\toprule
\textbf{Model} & \textbf{\#Text}  & \textbf{LLM Params} & \textbf{Frames} & \textbf{Short} & \textbf{Medium} & \textbf{Long} & \textbf{Overall} & \textbf{Gain} \\
  \midrule
    \multicolumn{9}{c}{\textbf{\textit{Proprietary LVLMs}}} \\
\midrule
\rowcolor{gray!20} GPT-4o \cite{gpt4o} & - & - & 384 & 80.0 & 70.3 & 65.3 & 71.9 & - \\
\rowcolor{gray!20} Gemini-1.5-Pro \cite{gemini} & - & - & 0.5 fps & 81.7 & 74.3 & 67.4 & 75.0 & - \\
  \midrule
      \multicolumn{9}{c}{\textbf{\textit{Open-Source LVLMs}}} \\
      \midrule
Video-LLaVA \cite{videollava} & - & 7B & 8 & 44.6 & 38.3 & 35.8 & 39.6 & - \\ 
\rowcolor{cyan!10} Video-LLaVA + TV-RAG &2.0K & 7B & 8 & 49.7 & 44.3 & 42.6 & 44.1 & \textcolor{red}{+4.5} \\ 
LLaVA-NeXT-Video \cite{llavanextvideo} & - & 7B & 16 & 49.4 & 43.0 & 36.7 & 43.0 & - \\ 
\rowcolor{cyan!10} LLaVA-NeXT-Video + TV-RAG & 2.0K & 7B & 16 &  70.4 & 53.7  & 54.1 & 58.8 & \textcolor{red}{+15.8} \\ 
LongVA \cite{longva} & - & 7B & 32 & 60.9 & 49.3 & 44.0 & 51.4 & - \\ 
\rowcolor{cyan!10} LongVA + TV-RAG & 1.8K & 7B & 32 & 66.2& 62.1 &58.7  &61.7& \textcolor{red}{+10.3} \\ 
Long-LLaVA \cite{longllava} & - & 7B & 32 & 60.3 & 51.4 & 44.1 & 52.0 & - \\ 
\rowcolor{cyan!10} Long-LLaVA + TV-RAG & 1.9K & 7B & 32 & 66.4 & 60.2 & 59.8 & 62.1 & \textcolor{red}{+10.1} \\ 
Qwen2-VL \cite{qwen2vl} & - & 72B & 32 & 75.0 & 63.3 & 56.3 & 64.9 & - \\ 
\rowcolor{cyan!10} Qwen2-VL + TV-RAG & 2.1K & 72B & 32 & 76.3 & 70.5 & 72.1 & 74.9 & \textcolor{red}{+10.0}  \\
LLaVA-Video \cite{llavavideo} & - & 72B & 32 & 78.0 & 63.7 & 59.6 & 67.1 & -  \\ 
\rowcolor{cyan!10} LLaVA-Video + TV-RAG & 2.1K & 72B & 32 & \textbf{82.3} & \textbf{73.1} & \textbf{73.8} & \textbf{76.5} & \textcolor{red}{\textbf{+9.4}}\\

  \bottomrule
\end{tabular}
\caption{Performance evaluation on the Video-MME \cite{videomme} benchmark.The {\#Text} column logs the average volume of tokens that \mbox{TV\textendash RAG} appends for each sample.    
Most strikingly, the 72 B LLaVA-Video model~\cite{llavavideo} outfitted with our adapter edges past the commercial Gemini-1.5-Pro~\cite{gemini}.  The baseline results are from \cite{luo2024video}.
All open-source baselines and their \mbox{TV\textendash RAG} variants were re-benchmarked.
}
\label{tab_compare}
\end{table*}

\subsection{Datasets}
We gauge the effectiveness of our framework on three publicly accepted long-video testbeds.  
Video-MME \cite{videomme} contains real-world clips that run anywhere from 11 s to roughly one hour and probe fine-grained understanding of everyday scenarios. MLVU \cite{mlvu} spans nine evaluation tasks sourced from videos lasting 3 min to 2 h (mean length about 12 min), challenging models to reason across extended temporal ranges. Finally, LongVideoBench \cite{lvb} targets multimodal retrieval and reasoning in lengthy footage, offering 6,678 human-written multiple-choice questions distributed over 17 thematic categories.

\subsection{Experimental Implementation}
All experiments were executed on NVIDIA A100 GPUs.  At the first stage, we prune the LVLM's detection prompts, keeping only concrete, CLIP-responsive objects and discarding abstract terms.  
For following retrieval, we fix both the FAISS and CLIP acceptance thresholds to \(0.3\); FAISS scores are computed with the \texttt{IndexFlatIP} backend~\cite{faiss}.  
Our ablation suite is built around Long-LLaVA 7B \cite{longllava}, whose extended context window and modest footprint make it ideal for probing how RAG similarity cut-offs and frame-sampling rates affect performance.

\subsection{Main Results}

\noindent\textbf{Video-MME}. 
To conduct a fair test of \mbox{TV-RAG}, we fixed every candidate LVLM to the same 32-frame input budget, an especially practical ceiling for the 72B-parameter models and a convenient checkpoint for their 7B counterparts. 
Our benchmark therefore covers six systems: four open 7B backbones (Video-LLaVA~\cite{videollava}, LLaVA-NeXT-Video~\cite{llavanextvideo}, LongVA~\cite{longva}, Long-LLaVA~\cite{longllava}) and two 72B giants (Qwen2-VL~\cite{qwen2vl}, LLaVA-Video~\cite{llavavideo}). 
As reported in Table~\ref{tab_compare}, attaching \mbox{TV-RAG} to the 72B pair lifts their scores beyond the proprietary Gemini-1.5-Pro baseline~\cite{gemini}. 
Averaged over all six backbones, the pipeline delivers a significant gain, with the most dramatic jumps on extended footage. 
The boost stems from injecting $\sim$14 additional keyframes (about 2\,000 tokens in total, given the typical 144-token payload per frame), which compensates for the weak visual grounding of LVLMs that were largely pre-trained on text. 
These extra, text-rich frames act as semantic anchors and enable sharper comprehension of complex video content.

\noindent\textbf{MLVU}. 
Table~\ref{tab_mlvu} contrasts leading LVLMs on the benchmark's multiple-choice split. The vanilla 7B LLaVA-Video line scores 70.8, but once TV-RAG is attached, the figure climbs to 72.6, an absolute gain of $+1.8$ that propels the small model ahead of every open-source rival under 70B, including the 32B Oryx-1.5 (72.3). A similar but smaller gain is observed for the 72B backbone, improving from 73.1 to 73.4 and further widening the gap with the proprietary GPT-4o baseline (64.6). These improvements, achieved with the same 64-frame budget, underline TV-RAG's ability to supply complementary textual cues that sharpen video understanding, especially for lighter models that still have headroom to benefit from richer context.

\begin{table}[]
\centering 
\setlength{\tabcolsep}{1.2mm} 
\renewcommand\arraystretch{0.95} 
\begin{tabular}{lcc|c}
\toprule

\textbf{Model} & \textbf{\#Params}  & \textbf{Frames} & \textbf{Overall}   \\ \midrule
\multicolumn{4}{c}{\textbf{\textit{Proprietary LVLMs}}} \\ \midrule
\rowcolor{gray!20} GPT-4o \cite{gpt4o} & - & 0.5 fps & 64.6   \\   \midrule
\multicolumn{4}{c}{\textbf{\textit{Open-Source LVLMs}}} \\   \midrule
Video-CCAM \cite{videoccam} & 14B & 96 & 63.1  \\
Video-XL \cite{videoxl} & 7B & 256 & 64.9  \\
Aria \cite{aria} & 25.3B & 256 & 70.6   \\
LLaVA-Video* \cite{llavavideo} & 7B & 64 & 70.8 \\
Oryx-1.5 \cite{oryx} & 32B & 128 & 72.3 \\
LLaVA-Video* \cite{llavavideo} & 72B & 64 & \underline{73.1} \\ \midrule
\rowcolor{cyan!10} LLaVA-Video + TV-RAG & 7B & 64 & 72.6 \\
\rowcolor{cyan!10} LLaVA-Video + TV-RAG & 72B & 64 & \textbf{73.4} \\
\bottomrule
\end{tabular}
\caption{The overall performance in the multiple-choice task of the MLVU \cite{mlvu} benchmark. * donates the results of our replication.}
\label{tab_mlvu}
\end{table}

\noindent\textbf{LongVideoBench}. 
We next gauge the influence of TV\textendash RAG on both 7B and 72B LLaVA\textendash Video backbones via the LongVideoBench suite~\cite{lvb}.  For clarity, we adopt the plain 64–frame feed and omit the benchmark's optional interleaved prompt.  Table~\ref{tab_lvb} reveals that the 72B configuration, once augmented with our retrieval layer, climbs to the runner-up position on the validation leaderboard: its score edges past Gemini-1.5-Pro~\cite{gemini} by $1.6\%$ and trails GPT-4o~\cite{gpt4o} by merely $1.1\%$.  The lighter 7B model also profits, posting a $+2.2\%$ uplift under the same experimental regime.  These results underscore \mbox{TV\textendash RAG}’s capacity to boost long-video reasoning across model scales without altering input structure.

\begin{table}[]
\centering 
\setlength{\tabcolsep}{0.8mm} 
\renewcommand\arraystretch{0.95} 
\begin{tabular}{lcc|c}
\toprule

\textbf{Model} & \textbf{\#Params}  & \textbf{Frames} & \textbf{Overall}   \\ \midrule
\multicolumn{4}{c}{\textbf{\textit{Proprietary LVLMs}}} \\ \midrule
\rowcolor{gray!20} Gemini-1.5-Pro \cite{gemini} & - & 256 & 64.0   \\ 
\rowcolor{gray!20} GPT-4o \cite{gpt4o} & - & 256 & 66.7   \\ 
 \midrule
\multicolumn{4}{c}{\textbf{\textit{Open-Source LVLMs}}} \\   \midrule
VideoChat2-Mistral \cite{videochat} & 7B & 8 & 39.3  \\
ShareGPT4Video \cite{chen2024sharegpt4video} & 7B & 8  & 39.7   \\
LLaVA-Next-Mistral \cite{llavanext} & 7B & 8 & 49.1   \\ 
PLLaVA \cite{pllava} & 34B & 16 & 53.2   \\ 
LLaVA-Video \cite{llavavideo} & 7B & 64 & 56.6   \\ 
LLaVA-Video \cite{llavavideo} & 72B & 64 & \underline{61.9}  \\ \midrule
\rowcolor{cyan!10} LLaVA-Video + TV-RAG &  7B & 64 & 58.8 \\
\rowcolor{cyan!10} LLaVA-Video + TV-RAG & 72B & 64 & \textbf{65.6} \\

\bottomrule
\end{tabular}
\caption{The overall performance on the validation set of LongVideoBench \cite{lvb}. }
\label{tab_lvb}
\end{table}

\subsection{Ablation Studies}

\begin{figure}[t!]
  \centering 
  \includegraphics[width=1.0\linewidth]{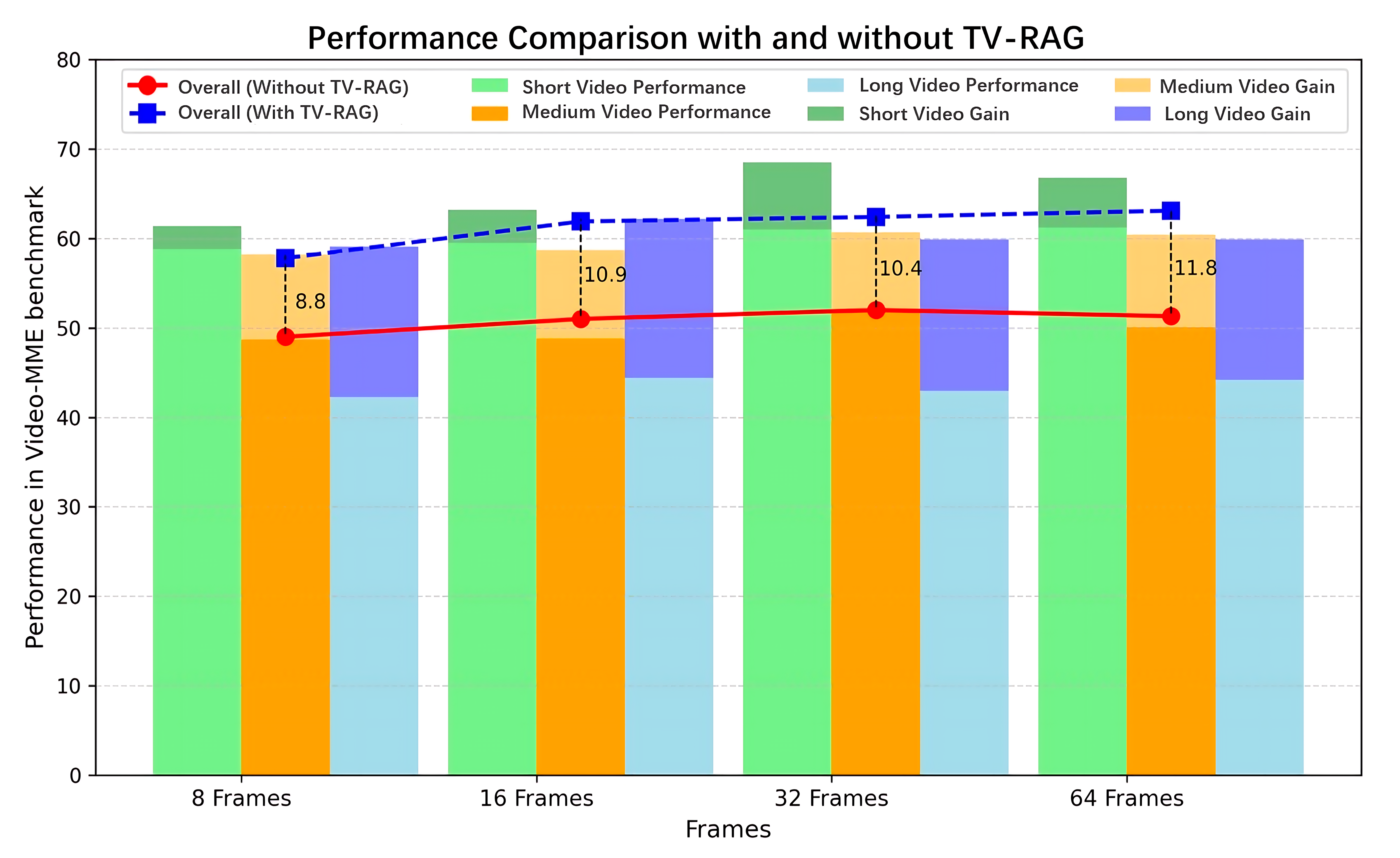}
  \caption{Performance with different sampling frames rate on Video-MME \cite{videomme} when using Long-LLaVA-7B \cite{longllava} as the LVLM.}
  \label{fig_abs_frame} 
\end{figure}

We probed the sensitivity of \mbox{TV\textendash RAG} to the volume of visual evidence by running the Long\textendash LLaVA\textendash 7B backbone~\cite{longllava} under four budgets: 8, 16, 32, and 64 sampled frames. 
Figure~\ref{fig_abs_frame} charts the accuracies and reveals two clear trends.  
First, the retrieval layer delivers a measurable lift at every budget; second, the gap widens as more frames are fed in, with the biggest jump on lengthy clips.  
In the vanilla baseline (no retrieval), performance tops out at the 32-frame mark, which demonstrates the effectiveness of our model.

We isolate each textual cue produced by the TV-RAG retriever, including semantic-entropy weighting (SE), temporal-window gating (TW), OCR strings, ASR transcripts, and the context-enrichment layer, and feed successive combinations to Long\textendash LLaVA\textendash 7B~\cite{longllava}. 
Performance on Video-MME~\cite{videomme} (Table~\ref{abl_comp}) climbs steadily with every new source, with ASR and OCR delivering the significant jump across short and long clips alike, confirming the unique value of ASR and OCR transcripts. 

SE (semantic-entropy) and TW (temporal-window) act as quality filters, discarding semantically weak or temporally mismatched snippets so that only the most relevant text aligns with each frame sequence. 
Taken together, these findings affirm the rationale behind \mbox{TV-RAG}: multimodal retrieval, temporal alignment, and adaptive weighting converge to produce more faithful and grounded video-language reasoning.

\begin{figure}[h!]
  \centering 
  \includegraphics[width=1.0\linewidth]{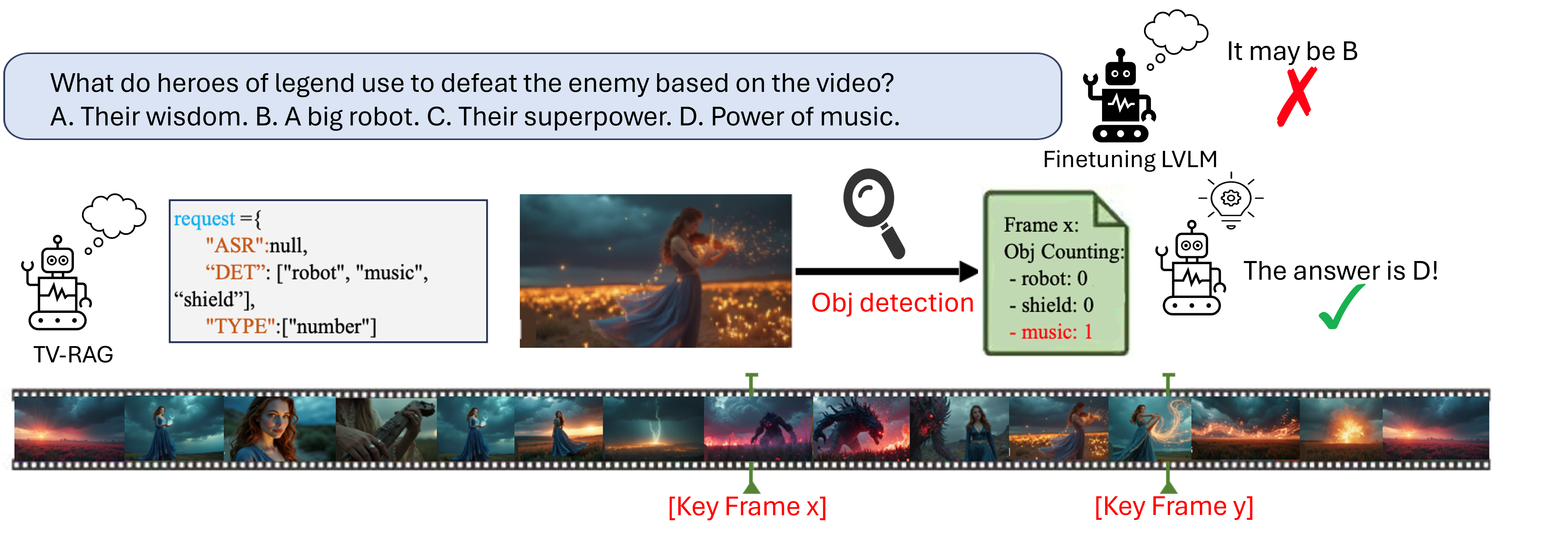}
  \caption{Qualitative result shown in Video-MME \cite{videomme} benchmark when applying TV-RAG with LLaVA-Video \cite{llavavideo}.}
  \label{fig_exam} 
\end{figure}

\begin{figure*}[t]
  \centering 
  \includegraphics[width=1.0\linewidth]{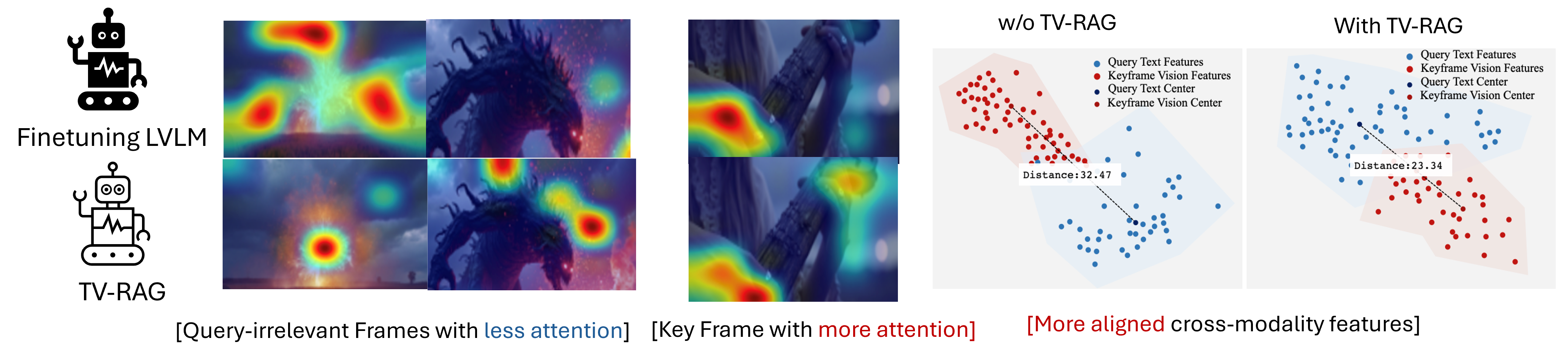}
  \caption{Grad-CAM heatmaps of the final hidden state, alongside t-SNE projections of the user's query and keyframe features, are visualized for the example in Figure \ref{fig_exam}. The combined visualization makes clear that the supplementary texts retrieved by \mbox{TV-RAG} tighten the vision–language link: they redirect the network's attention toward the frames most pertinent to the question, thereby boosting both answer precision and contextual fidelity.}
  \label{fig_tsne} 
\end{figure*}

\begin{table}
  \centering
  \resizebox{0.48\textwidth}{!}{ 
  \begin{tabular}{ccccc|ccc|c}
  \hline
  \textbf{SE} & \textbf{TW} & \textbf{OCR} & \textbf{ASR} & \textbf{Context} & \textbf{Short} & \textbf{Medium} & \textbf{Long} & \textbf{Overall} \\
  \hline
  $\times$ & \checkmark & \checkmark & \checkmark & \checkmark & 65.3 & 56.5 & 49.1 & 57.0 \\
  \checkmark & $\times$ & \checkmark & \checkmark & \checkmark & 64.5 & 52.7 & 50.3 & 55.8 \\
  \checkmark & \checkmark & $\times$ & \checkmark & \checkmark & 68.9 & 57.8 & 55.0 & 60.6 \\
  \checkmark & \checkmark & \checkmark & $\times$ & \checkmark & 68.3 & 59.4 & 57.8 & 61.8 \\
  \checkmark & \checkmark & \checkmark & \checkmark & $\times$ & 65.4 & 55.6 & 56.4 & 59.1 \\
  \rowcolor{cyan!10} \checkmark & \checkmark & \checkmark & \checkmark & \checkmark & \textbf{72.4} & \textbf{64.1} & \textbf{59.3} & \textbf{65.3} \\
  \hline
  \end{tabular}
  }
  \caption{Results for combinations of different extracted texts in Video-MME using LLaVA-Video-7B as the LVLM.}
  \label{abl_comp}
  \end{table}

Table~\ref{abl_thres} reports how the similarity cutoff $\tau$ shapes both efficiency and accuracy when Long-LLaVA 7B answers the benchmark's multiple-choice questions.  
Lower bars ($\tau{=}0$--$0.1$) flood the prompt with more auxiliary tokens, pushing the inference time to 30-40 s but only edge the overall score to 63-64.  
Raising the bar gradually prunes redundant text and accelerates decoding; at $\tau{=}0.3$ the model processes 1.9 K tokens in 13 s yet still logs 61.6 overall, while posting the single best medium-clip accuracy (61.0).  
Beyond this point, the gains erode: $\tau{=}0.5$ and $\tau{=}1.0$ shrink the token load to $\le0.3$ K and time to $\le9$ s, but overall correctness drops sharply to 56.7 and 52.5, respectively.  
Hence $\tau{=}0.3$ offers the most balanced trade-off—halving latency relative to the lowest cutoffs, staying within typical context limits (\,$\sim$1.9 K extra tokens), and preserving strong performance across clip lengths.  
For LVLMs with tighter windows, a softer cutoff such as $\tau{=}0.2$ may be preferred, still keeping the token budget under 3 K while avoiding the steep accuracy loss seen at stricter thresholds.

 To study the effectiveness of our model, we compare it with Video-RAG \cite{luo2024video}. As shown in Table~\ref{other-model-plug}, our method demonstrates a clear performance boost over Video-RAG across all video lengths. This improvement is observed in both the Long-Video and LLaVA-Video models, with particularly noticeable gains in handling shorter and longer videos. Overall, our model contributes to better model performance by effectively addressing the challenges posed by different video durations. The observed gains can be attributed to our framework's ability to integrate temporal semantics more adaptively. 

\begin{table}[H]
\centering 
\resizebox{0.48\textwidth}{!}{
\setlength{\tabcolsep}{1.3mm} 
\renewcommand\arraystretch{0.95} 
\begin{tabular}{c|cc|cccc} 
\toprule
\textbf{Model} & \textbf{Params} & \textbf{Frames} & \textbf{Short} & \textbf{Medium}  & \textbf{Long} & \textbf{Overall}   \\ \midrule
X=LongVA &7B &32 &60.9& 49.3& 44.0& 51.4\\
X+Video-RAG& 7B& 32& 65.4& 59.1& 55.7& 60.1\\
\rowcolor{cyan!10} X+  TV-RAG& 7B &32 &66.2& 62.1 &58.7  &61.7\\ 
 \midrule
 X=LLaVA-Video & 72B &32 &78.0& 63.7& 59.6& 67.1 \\
 X+Video-RAG&  72B &32& 81.1& 72.9& 73.1 &75.7\\
\rowcolor{cyan!10} X+ TV-RAG &72B& 32 &82.3& 73.1& 73.8& 76.5\\
\bottomrule
\end{tabular}
}
\caption{Performance with other model on Video-MME.}
\label{other-model-plug}
\end{table}

\begin{table}[]
\centering 
\setlength{\tabcolsep}{1.3mm} 
\renewcommand\arraystretch{0.95} 
\begin{tabular}{c|cc|ccc|c}
\toprule
\textbf{$\tau$} & \textbf{\#Token} & \textbf{Time} & \textbf{Short} & \textbf{Medium}  & \textbf{Long} & \textbf{Overall}   \\ \midrule
0.0 & 3.6K & 39s & 66.3 & 58.4 & 57.1 & 63.4 \\
0.1 & 3.4K & 33s & 68.7 & 57.3 & 59.1 & 62.9 \\
0.2 & 2.7K & 18s & 66.4 & 60.2 & 58.2 & 60.5 \\
\rowcolor{cyan!10} 0.3 & 1.9K & 13s & 67.4 & \textbf{61.0} & 58.8 & 61.6 \\
0.4 & 0.8K & 10s & 64.6 & 58.8 & 58.5 & 60.3 \\
0.5 & 0.3K & 9s & 63.4 & 54.5 & 50.1 & 56.7 \\
1.0 & 0.0K & 8s & 60.4 & 51.3 & 44.7 & 52.5 \\
\bottomrule
\end{tabular}
\caption{Performance with different thresholds of retrieval on Video-MME \cite{videomme} when using Long-LLaVA 7B \cite{longllava} as the LVLM. \textbf{\#Token} and \textbf{Time} denote the total token number of the extracted texts and the average inference time per question, respectively. }
\label{abl_thres}
\end{table}

\subsection{More Studies}

We present more expriments for Video-MME \cite{videomme} in Figure \ref{fig_exam} and Figure \ref{fig_tsne} to illustrate the effectiveness of our approach. As shown in the figures, integrating external tools with LLaVA-Video to process and retrieve extracted textual information from videos significantly enhances the model's ability to mitigate visual hallucinations. This integration leads to more accurate and contextually grounded responses to user queries by reinforcing the alignment between visual content and textual understanding.

Furthermore, the Grad-CAM \cite{gradcam} and t-SNE \cite{tsne} visualization results provide additional evidence supporting the impact of TV-RAG in strengthening cross-modal alignment within the LVLM. Specifically, the Grad-CAM results highlight improved attention to semantically relevant regions in the video frames, while the t-SNE analysis reveals a more structured representation space, indicating a refined relationship between visual and textual modalities. These findings collectively underscore the effectiveness of our approach in enhancing the robustness and interpretability of LVLM-based video understanding.

In summary, the qualitative examples not only highlight the correctness of predictions but also provide insight into how the model reasons through retrieved information.

\section{Conclusion}
\label{sec:con}

We propose TV-RAG, a training-free adaptor that upgrades long-video reasoning in large vision-language models.  The scheme fuses two key ideas: (i) a sliding temporal window that aligns retrieved cues with the correct time span and (ii) an entropy-based score that favours semantically rich segments, together compensating for the narrow temporal receptive field of standard LVLMs and for their difficulty in tracing subtle scene-level shifts.  Contrary to retrievers that rely purely on text similarity or pipelines that demand full model fine-tuning, \mbox{TV-RAG} clips onto any existing backbone as a light, plug-and-play module.  
Extensive trials on Video-MME, MLVU, and LongVideoBench confirm marked gains, allowing the patched models to overtake contemporary leaders such as Gemini-1.5-Pro and GPT-4o in some metrics.   
The future work targets finer temporal gating and adaptive entropy scaling to further strengthen cross-modal alignment and deep contextual reasoning.
In future work, we will extend TV-RAG to dynamically adjust its temporal window size based on content uncertainty and explore curriculum-style retrieval schedules to better adapt to ultra-long, open-world video streams. We also plan to generalise the framework to audio-visual co-reasoning and egocentric or 360° footage, broadening its applicability across emerging multimodal benchmarks.

\section{Acknowledgments}
This work was supported by the Science and Technology Innovation 2030-Key Project under Grant 2021ZD0201404.

\bibliographystyle{ACM-Reference-Format}
\balance
\bibliography{main}
\end{document}